\documentclass[english]{lni}
\usepackage{booktabs}
\usepackage{graphicx}
\usepackage{float}
\usepackage{subcaption}
\usepackage{multirow}
\begin{document} 
\title[Detection and Simulation of Urban Heat Islands Using a Fine-Tuned Geospatial Foundation Model]{Detection and Simulation of Urban Heat Islands Using a Fine-Tuned Geospatial Foundation Model}
 \author[1]{David Kreismann}{davidkreismann@gmx.de}{0009-0000-0626-7409}
 \affil[1]{Baden-Wuerttemberg Cooperative State University\\Paulinenstraße 50\\70178 Stuttgart\\Germany}
\maketitle

\begin{abstract}
As urbanization and climate change progress, urban heat island effects are becoming more frequent and severe. To formulate effective mitigation plans, cities require detailed air temperature data. However, predictive analytics methods based on conventional machine learning models and limited data infrastructure often provide inaccurate predictions, especially in underserved areas. In this context, geospatial foundation models trained on unstructured global data demonstrate strong generalization and require minimal fine-tuning, offering an alternative for predictions where traditional approaches are limited. This study fine-tunes a geospatial foundation model to predict urban land surface temperatures under future climate scenarios and explores its response to land cover changes using simulated vegetation strategies. The fine-tuned model achieved pixel-wise downscaling errors below 1.74 °C and aligned with ground truth patterns, demonstrating an extrapolation capacity up to 3.62 °C.
\end{abstract}

\section{Introduction}
As cities grow, increasing building and population densities drive land use changes that impact the local energy balance and shape the microclimate and thermal conditions of urban areas \citep{nurwandaPredictionCityExpansion2020}.
With over 70\% of the global population projected to live in cities by 2050, the urban heat island (UHI) effect has become a major concern \citep{LIN2023109910}.
Factors such as heat-retaining materials, vegetation loss, dense structures, and human activity can cause cities to be over 5$^\circ$C warmer than their surrounding areas \citep{book}. These temperature increases lead to greater energy consumption, more heat-related health issues, and declining air and water quality \cite{tanUrbanHeatIsland2010}.

One major challenge in addressing UHI impacts is the lack of high-resolution, timely air temperature data, which is essential for early warnings and effective heat risk mitigation plans \cite{kousisIntraurbanMicroclimateInvestigation2021}. Current forecasting approaches often require large amounts of input data, specialized expertise, and significant computational resources, making real-time forecasting at fine scales largely inaccessible \cite{powersWeatherResearchForecasting2017}. 
Data limitations from sparse stations and satellite revisit intervals further constrain accuracy \cite{varentsovMachineLearningSimulation2023}. 
While artificial intelligence (AI)-based models offer increased flexibility, they are often limited by their dependence on large labeled datasets and poor generalization across domains \cite{farahani2020briefreviewdomainadaptation}.

In response to these limitations, recent developments in geospatial foundation models (GFMs), trained on vast, unstructured global datasets, have demonstrated strong generalization across spatial resolutions and geographic regions with minimal fine-tuning, making them a promising alternative for urban climate applications \cite{mai2023opportunitieschallengesfoundationmodels}. 
Building on this potential, the study explores the use of a GFM to simulate high-resolution land surface temperatures (LST) in urban environments. Specifically, it focuses on forecasting UHI effects in Brașov, Romania, and evaluating the model’s ability to predict temperature responses under varying climate and vegetation scenarios. The goal is to show the potential of GFMs for improving urban heat mitigation and climate adaptation.

\section{Related work}
\textcite{kimMaximumUrbanHeat2002} presented one of the first ML based approaches for UHI prediction. The study focused on Seoul, Korea, using meteorological data from urban and rural stations collected between 1973 and 1996. To model these patterns, they used multiple linear regression based on four predictors, previous day UHI intensity, wind speed, cloudiness, and relative humidity. Although such modeling techniques can be effective in certain contexts, they often struggle to generalize in complex urban environments \cite{varentsovMachineLearningSimulation2023}.

Building on these foundations, recent approaches apply more advanced methods supported by richer datasets. For example, \textcite{baduguPredictingLandSurface2024} used a long short-term memory model with two decades of satellite data from Tiruchirappalli, India, integrating variables such as aerosol optical depth, vegetation index, and elevation. Similarly, \textcite{guptaUrbanLandSurface2024} applied neural networks to forecast LST in Kolkata, using land cover, elevation, and historical temperature patterns. In both cases, the studies effectively captured spatial thermal patterns in rapidly urbanizing areas.

However, despite their potential, conventional machine learning approaches face several challenges. Many models lack physical interpretability and require extensive city-specific calibration, which limits their generalizability \cite{varentsovMachineLearningSimulation2023}. Deep learning methods, though powerful, are further constrained by the need for large volumes of annotated training data, a requirement that remains difficult to scale, even with conventional labeling techniques \cite{varentsovMachineLearningSimulation2023}.

To address these limitations, recent approaches have explored the use of GFMs.
Unlike traditional ML models, which often struggle with variations in urban structures, GFMs offer more adaptable and transferable solutions for urban heat mapping across diverse cities \cite{mai2023opportunitieschallengesfoundationmodels}.
In this context, \textcite{jakubik2023foundationmodelsgeneralistgeospatial} introduced Prithvi, a transformer-based GFM model pretrained via masked autoencoding on Harmonized Landsat–Sentinel-2 imagery. The model splits input images into patches, randomly masks a subset, and reconstructs them using a Vision Transformer (ViT) encoder–decoder architecture. It is trained by minimizing the mean squared error over the masked regions, which promotes the learning of semantically meaningful representations. Prithvi can be fine-tuned for classification or pixel-wise regression tasks, making it well suited for UHI prediction \cite{jakubik2023foundationmodelsgeneralistgeospatial}.

To the best of my knowledge, the study by \textcite{10641750} is the only published work to date that applies a GFM for UHI prediction. Focusing on Johannesburg, South Africa, the authors fine-tuned the Prithvi model to estimate 2-meter air temperature (T2M) at 1 km resolution, achieving a mean absolute error (MAE) below 1.5\,${}^\circ$C. The model combines HLS imagery with ERA5 T2M data and is trained on observations from global cities spanning diverse hydroclimatic zones from 2013 to 2023. A SWIN transformer backbone captures spatial patterns at multiple scales through local and global self-attention, and a unified regression head integrates heterogeneous inputs for temperature prediction \cite{10641750}.

\section{Methodology}
GFMs have shown strong performance in geospatial tasks, with promising generalization across space and time, making them suitable for UHI prediction under growing urbanization and climate extremes \cite{jakubik2023foundationmodelsgeneralistgeospatial}. However, applications involving simulation methods such as inpainting or forecasting under future climate scenarios are still rare, with only preliminary work reported by \textcite{10641750}.

To address this gap, this study evaluates a fine-tuned geospatial foundation model for simulating UHI effects under varying climate and land cover scenarios. Building on the model introduced by \textcite{10641750}, the methodology extends the approach to forecast UHI impacts under future Representative Concentration Pathway (RCP) scenarios and to simulate mitigation strategies through pixel-wise inpainting and spectral index modifications. The approach includes both zero-shot and fine-tuned evaluations, focusing on the model’s ability to extrapolate beyond its training distribution. To test this, the dataset is split by temperature, with higher-temperature instances held out to assess generalization to extreme heat. The model is applied to the study area described in Section~\ref{section:StudyRegion} and benchmarked against satellite-derived ground truth data.

\subsection{Study region}\label{section:StudyRegion}
\begin{figure}[H]
    \centering
    \resizebox{0.7\textwidth}{!}{%
      \makebox[\textwidth][c]{%
        \begin{minipage}[t]{0.58\textwidth}
          \begin{subfigure}[t]{\textwidth}
            \includegraphics[width=\textwidth]{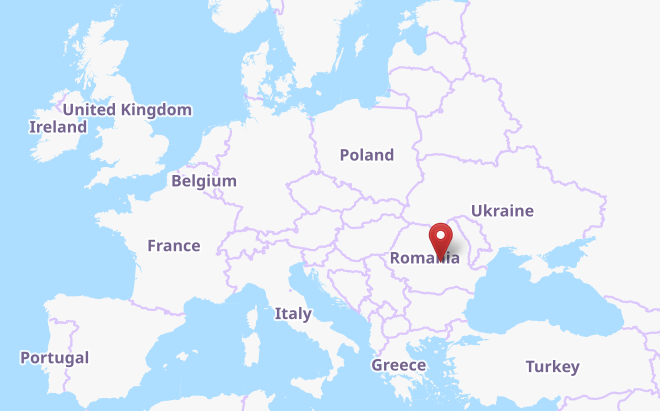}
            \caption{Study area of the Brașov metropolitan region location within Europe.}
            \label{fig:europe}
          \end{subfigure}
  
          \vspace{1em}
  
          \begin{subfigure}[t]{0.49\textwidth}
            \includegraphics[width=\textwidth]{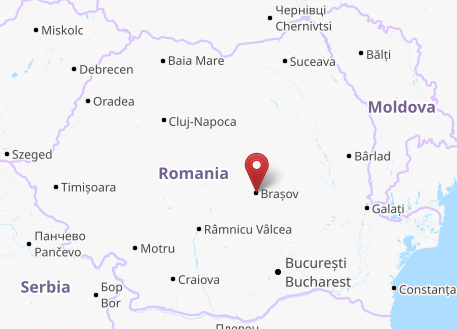}
            \caption{Study area of the Brașov metropolitan region location within Romania.}
            \label{fig:boundary}
          \end{subfigure}\hfill
          \begin{subfigure}[t]{0.49\textwidth}
            \includegraphics[width=\textwidth]{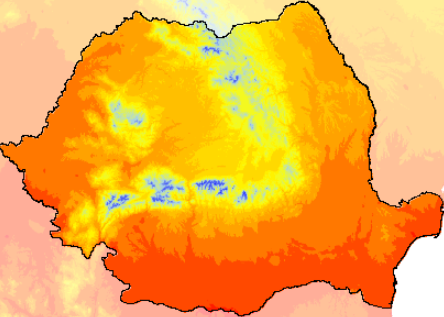}
            \caption{Average annual temperature across Romania (dark red \(\approx12\,^\circ\mathrm{C}\), dark blue \(\approx-2\,^\circ\mathrm{C}\)).}
            \label{fig:terrain}
          \end{subfigure}
        \end{minipage}%
        \hspace{0.02\textwidth}
  
        \begin{minipage}[c]{0.38\textwidth}
          \begin{subfigure}[c]{\textwidth}
            \includegraphics[width=\textwidth]{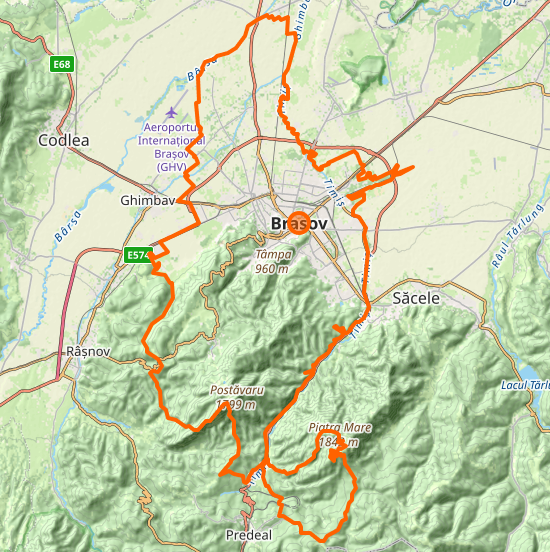}
            \caption{Topographic representation of Brașov City showing the Brașov metropolitan area boundary (orange line).}
            \label{fig:landcover}
          \end{subfigure}
        \end{minipage}%
      }
    }
  
    \caption[Overview of the Brașov city study area]{Overview of the Brașov city study area. Maps are derived from geospatial data provided by Haklay, Weber (2008) \cite{article255} and Robert, Hijmans (2022) \cite{GADM}.}
    \label{fig:studyarea}
  \end{figure}

The study concentrates on a single city to allow for a more focused evaluation of the model's performance. Although GFMs are designed to generalize across multiple areas, Brașov, Romania was selected as it had not been included in the model’s training data, and its urban and environmental characteristics offer a challenging setting for UHI analysis \cite{mihaiNatureBasedSolutionsClimateResilient2024}.
Brașov is a historic city in central Romania, located approximately 161 km north of Bucharest, situated within the Carpathian Mountains at an average elevation of 625 meters. Figure~\ref{fig:studyarea} shows Brașov’s topographic setting, characterized by valley floors, hills, and surrounding mountains that create notable microclimatic variation \cite{bogdanHeritageBuildingPreservation2022}. 

Despite its forested surroundings, Brașov is among Romania’s most polluted cities, reflecting the need for green infrastructure that improves cooling, air quality, and water regulation. While parks and green areas play an important role in the urban environment, many are based on outdated design principles and feature low plant diversity \cite{mihaiNatureBasedSolutionsClimateResilient2024}. As temperatures rise and maintenance demands grow, the need for climate-informed planning and resilient urban design becomes increasingly urgent. These challenges underscore the potential of AI-based approaches to strengthen green infrastructure and support data-driven approaches to UHI mitigation.

\subsection{Datasets}
This study relies on multiple open-access geospatial datasets for the period from 2013 to 2023, including Landsat 8 imagery \cite{royLandsat8ScienceProduct2014}, ERA5 reanalysis data \cite{MunozSabater2019}, CORDEX climate model outputs \cite{C3S_CORDEX2019}, and ESRI land use/land cover (LULC) maps \cite{9553499}. These datasets were obtained from publicly available repositories, including the United States Geological Survey (USGS), the Copernicus Climate Data Store (CDS), and the ESRI Sentinel-2 Land Cover Explorer. In the following, each dataset is briefly summarized with respect to its role in this study.

\textbf{Landsat 8 satellite imagery}: 
Developed by NASA and USGS, Landsat 8 captures global imagery in visible, near-infrared, shortwave, and thermal infrared bands, with a 16-day revisit cycle and 30m resolution \cite{royLandsat8ScienceProduct2014}. 
In this study, Landsat 8 Bands 1 to 6, covering the visible, near-infrared, and shortwave infrared regions, are used as input for the LST analysis.

\textbf{ERA5-Land reanalysis data}:
ERA5-Land is a global high-resolution land reanalysis dataset produced by the ECMWF \cite{essd-13-4349-2021}. 
The dataset offers hourly near-surface climate variables at 9km resolution. This study uses the T2M variable to assess thermal conditions.

\textbf{EURO-CORDEX climate model data}:
EURO-CORDEX, offers high-resolution regional climate projections at 12.5~km resolution using downscaled CMIP5 data \cite{esd-12-1543-2021}. In this study, the three-hourly T2M variable is used to evaluate UHI patterns under three different RCP scenarios, which represent standardized greenhouse gas trajectories \cite{mossNextGenerationScenarios2010}. The included scenarios are RCP 8.5 (high emissions, ~1370 ppm CO\textsubscript{2} by 2100), RCP 4.5 (stabilization, ~650 ppm), and RCP 2.6 (strong mitigation, peak near 2020, <2 °C warming).

\textbf{ESRI land use/land cover (LULC) data}:
LULC represents how land is used (e.g., agriculture, urban areas) and its surface cover (e.g., forests, water). ESRI provides annual global maps at 10m resolution, produced with a deep learning segmentation model, supporting consistent monitoring of land cover change over time \cite{9553499}.

\section{Experiments}\label{section:experiments}
While the original GFM by \textcite{10641750} was fine-tuned for predicting T2M, this study employs an adapted version for LST, as available on Hugging Face \cite{bhamjeeHuggingFaceGranite}. LST, derived from satellite observations, captures radiative surface temperature and is highly sensitive to urban form and material composition, which are key factors for UHI modeling and climate informed planning \cite{wenComparativeStudyEffects2024}.

The model is fine-tuned using paired inputs from Landsat 8 imagery and ERA5-Land T2M reanalysis data, following the procedure described by \textcite{10641750}. Training is performed using TerraTorch \cite{gomes2025terratorch} on a system with 8 CPU cores, 128 GB RAM, and a single NVIDIA A100 GPU. LST ground truth values are derived using the split-window algorithm \cite{duPracticalSplitWindowAlgorithm2015}.
Two pretrained model variants are used: one trained on 28 cities and another on 51 cities, both covering diverse hydroclimatic zones from 2013 to 2023.
Fine-tuning is conducted using the AdamW optimizer with a learning rate of $6\times10^{-5}$, GeLU activation, a batch size of 16, and up to 100 epochs. Model performance is evaluated using MAE.

As described, this study extends the base model to enable forecasting and simulation of mitigation scenarios. The full pipeline therefore consists of two stages, fine-tuning and inference, applied to the Brașov region. In the fine-tuning phase, the model is trained on Landsat 8 bands and ERA5 T2M data, using LST labels specific to Brașov. To assess extrapolation capacity, the dataset is split by temperature, with the top 10\% of values held out for testing to simulate generalization to extreme heat conditions.

In the inference stage, the fine-tuned model is used to predict LST across Brașov. The decoder outputs spatially detailed temperature maps based on the model’s learned representations. To enable future scenario analysis, the model is further tested using EURO-CORDEX T2M projections (RCP 2.6, 4.5, and 8.5) instead of ERA5 T2M.

To explore mitigation strategies, inpainting techniques are applied to simulate land cover changes. This involves modifying pixel values and vegetation-related indices such as the Normalized Difference Vegetation Index (NDVI), the Normalized Difference Built-up Index (NDBI), and the Normalized Difference Water Index (NDWI). These indices are defined as:

\begin{equation*}
NDVI = \frac{\rho_{NIR} - \rho_{Red}}{\rho_{NIR} + \rho_{Red}} \qquad
NDBI = \frac{\rho_{SWIR} - \rho_{NIR}}{\rho_{SWIR} + \rho_{NIR}} \qquad
NDWI = \frac{\rho_{\text{Green}} - \rho_{\text{NIR}}}{\rho_{\text{Green}} + \rho_{\text{NIR}}}
\end{equation*}

where $\rho_{NIR}$, $\rho_{Red}$, $\rho_{Green}$, and $\rho_{SWIR}$ represent sensorreflectance values from the near-infrared, red, green, and shortwave infrared bands of Landsat 8, respectively \cite{addasMachineLearningTechniques2023}. 

These indices are widely used to characterize land cover types and their thermal effects. NDVI serves as a proxy for vegetation density, with higher values indicating greater vegetation cover. NDBI reflects the extent of built-up surfaces, such as concrete and asphalt. NDWI indicates the presence of surface water or moisture content in vegetation. By adjusting these spectral bands in the satellite image of Brașov, the index values can be systematically increased or decreased to simulate different land cover scenarios. 

\section{Evaluation}
This section presents the experimental evaluation of the GFM and highlights key empirical findings.
Table~\ref{tab:model_comparison} presents the test set results for various GFM configurations. Both model variants were pretrained on global city data from 2013 to 2023, V1 on 28 cities and V2 on 51. Each model was then fine-tuned on data from Brașov, Romania. For consistency, all fine-tuned models used the same hyperparameters, detailed in Section~\ref{section:experiments}.
The \enquote{Baseline} reflects zero-shot performance without Brașov specific fine-tuning. The \enquote{Random Data Split} uses a 72/18/10 train/val/test split. In the \enquote{High Heat Scenario}, only temperature values up to the 90th percentile are used for training and validation, while the top 10 percent is reserved for testing.

\begin{table}[H]
  \centering
  \begin{tabular}{@{}l ccc ccc@{}}
    \toprule
    \multirow{2}{*}{Model Variants} & \multicolumn{3}{c}{\textbf{V1}} & \multicolumn{3}{c}{\textbf{V2}} \\
    \cmidrule(lr){2-4} \cmidrule(lr){5-7}
    & MAE & MSE & RMSE & MAE & MSE & RMSE \\
    \midrule
    Baseline                 & 1.95 & 7.25 & 2.69 & 2.81 & 12.94 & 3.60 \\
    Random Data Split        & 1.80 & \underline{6.26} & \underline{2.50} & \underline{1.77} & \textbf{5.80} & \textbf{2.41} \\
    High-Heat Scenario (90th) & 1.96 & 7.05 & 2.66 & \textbf{1.74} & 6.37 & 2.52 \\
    \bottomrule
  \end{tabular}
  \caption[Model comparison via MAE, MSE, RMSE]{Performance comparison of model variants using MAE, MSE, and RMSE. Bold indicates the best value and underline indicates the second-best.}
  \label{tab:model_comparison}
\end{table}
In summary, the \enquote{High Heat Scenario} model, which is fine-tuned on data below the 90th temperature percentile, performs comparably to the random split across all metrics and outperforms it in MAE for model V2. It consistently exceeds the Baseline, except for MAE in model V1. Across all setups and model variants, \enquote{High-Heat Scenario} differences remain below 0.60\,${}^\circ$C, indicating that GFMs can generalize beyond their training data and extrapolate to unseen temperature extremes. This underscores the effectiveness of transfer learning with pretrained foundation models.

In this experiment, the 90th percentile threshold was 23.29\,${}^\circ$C, with the GFM predicting up to 26.91\,${}^\circ$C and achieving an MAE of 1.74\,${}^\circ$C on model V2. Compared to the other model configurations, this demonstrates that the model successfully extrapolated 3.62,${}^\circ$C beyond its training range, effectively generalizing to conditions representing potential future warming in Brașov.

To better understand the model’s behavior, its performance was analyzed at the pixel level by LULC class. This assessment highlights where the model performs well or struggles across different land cover types. Table~\ref{tab:lulc_error_metrics} shows average error metrics per class for the fine-tuned V2 model under the \enquote{High-Heat Scenario}. 

\begin{table}[H]
  \centering
  \small 
  \begin{tabular}{@{}lccc c@{}}
    \toprule
    LULC Class
      & MAE 
      & RMSE 
      & MSE 
      & LULC Distribution (\%) \\
    \midrule
    Class 1 (Water)
      & 2.08 & 2.40 & 7.81 & 0.18 \\
    Class 2 (Trees)
      & \underline{1.73} & \underline{2.04} & \underline{6.10} & 8.41 \\
    Class 4 (Flooded Vegetation)
      & \textbf{1.31} & \textbf{1.39} & \textbf{3.23} & 0.01 \\
    Class 5 (Crops)
      & 1.85 & 2.22 & 6.07 & 53.80 \\
    Class 7 (Built Area)
      & 1.98 & 2.45 & 7.54 & 30.98 \\
    Class 8 (Bare Ground)
      & 2.15 & 2.34 & 8.98 & 0.09 \\
    Class 10 (Clouds)
      & 2.22 & 2.24 & 5.75 & 0.00 \\
    Class 11 (Rangeland)
      & 2.08 & 2.43 & 7.54 & 5.42 \\
    \bottomrule
  \end{tabular}
  \caption{Average error metrics and land cover share per LULC class, based on pixel-wise comparisons over the Brașov test set for the high-heat scenario using model V2. Bold indicates the best value and underline indicates the second-best.}
  \label{tab:lulc_error_metrics}
\end{table}

The GFM achieved its lowest MAE of 1.31\,${}^\circ$C in the \enquote{Flooded Vegetation} class and the highest error of 2.22\,${}^\circ$C in the \enquote{Clouds} class. Strong results were also observed for \enquote{Trees} (1.73\,${}^\circ$C) and \enquote{Built Area} (1.98\,${}^\circ$C), suggesting the model performs well in both vegetated and urban environments. The low error for \enquote{Flooded Vegetation} may be influenced by its minimal representation (0.01\,\%) in the dataset. In contrast, the robust performance in \enquote{Built Area}, which comprises 30.98\,\% of the data, highlights the model’s ability to generalize effectively across dominant land cover types, which supports its relevance for urban climate modeling.

\begin{figure}[H]
  \centering
  \captionsetup{list=yes} 
  \includegraphics[width=0.55\linewidth]{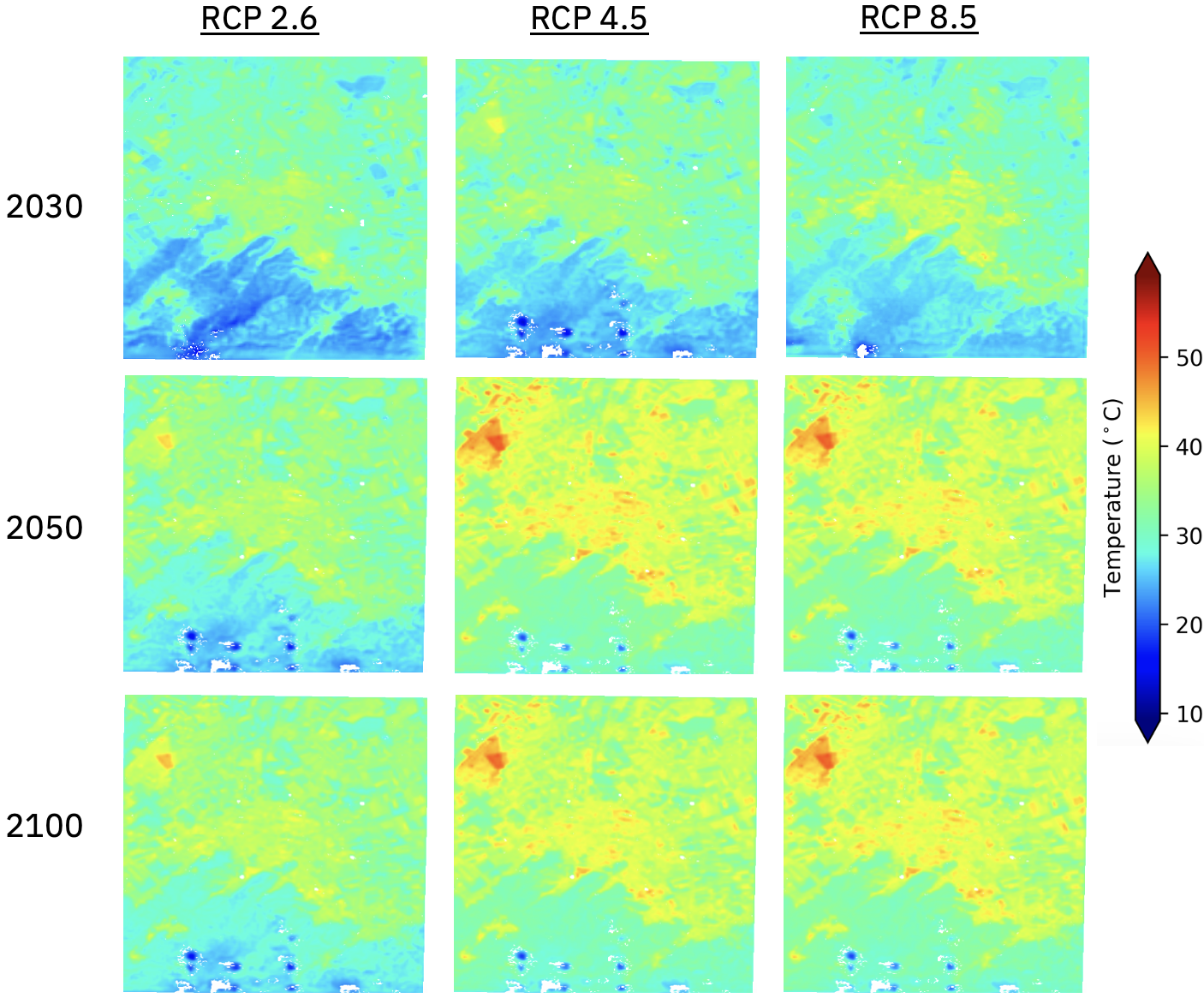}
  \caption{Projected UHI extent under RCP 2.6, 4.5, and 8.5 for 2030, 2050, and 2100 in Brașov, Romania. Temperature estimates are shown per pixel, highlighting spatial variations.}
  \label{fig:euro-cordex-forecasting}
\end{figure}

Given the model’s demonstrated extrapolation capacity of up to 3.62\,${}^\circ$C, it is important to further explore scenario-based forecasting under different RCP pathways. Assuming 3.62\,${}^\circ$C as the model’s upper limit, Figure~\ref{fig:euro-cordex-forecasting} shows projections for 2030, 2050, and 2100 based on EURO-CORDEX data using the global climate model CNRM-CERFACS-CM5 in combination with the regional model CNRM-ALADIN63. Results indicate a clear link between rising global temperatures and the expansion and intensification of UHI hotspots.

This underscores the value of climate scenario forecasting for understanding UHI dynamics and informing urban adaptation. Under the assumption of a constant landscape, it helps identify current and emerging hotspots and supports targeted mitigation planning.

The model is also evaluated for its ability to simulate landscape modification effects, supporting urban planning and climate mitigation strategies. This is done by modifying satellite imagery through pixel-wise swaps between land cover types to simulate alternative urban designs. Three scenarios are tested: (1) inserting forest pixels, (2) inserting urban pixels, and (3) retaining the original ground truth. The modified area, outlined in red, spans approximately $0.0195~\text{km}^2$ and intersects multiple LULC classes, forest, crops, and built-up areas, to assess the model’s sensitivity to surrounding context. Inference is run using the modified imagery to generate predicted LST maps. The pixel swap is performed within the red bounding box.
Vertical temperature profiles are plotted below each LST image. The blue line within the bounding box indicates a vertical sampling path through its center (top to bottom), corresponding to the left-to-right axis in the plots below. In these plots, blue segments represent unmodified regions, while red segments highlight temperature values within the modified area.

\begin{figure}[H]
  \centering
  \captionsetup{list=yes}
  \includegraphics[width=0.48\linewidth]{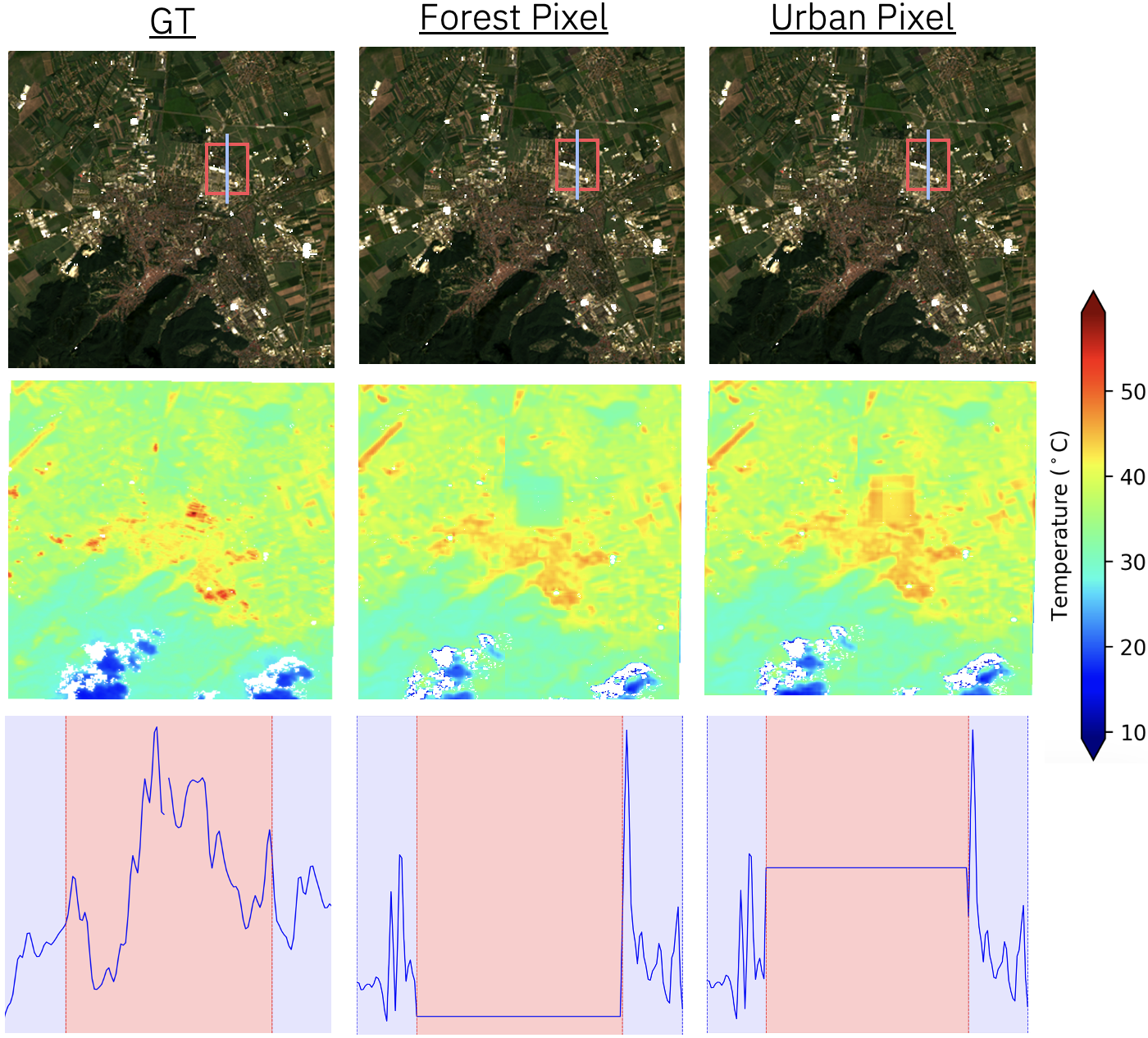}
  \caption[Effect of single-pixel replacement on inpainting results]{Comparison of pixel-wise inpainting against ground truth data, demonstrating the effect of swapping a pixel area with either a forest or urban pixel.}
  \label{fig:pixel-inpainting}
\end{figure}

The results in Figure~\ref{fig:pixel-inpainting} show that the model responds strongly to land cover changes. Inserting forest pixels reduces predicted LST, whereas urban pixels raise it. The line plots also reveal that temperatures just outside the modified area are affected, especially near the transition from blue to red, before gradually returning to baseline. This suggests that the GFM captures the spatial influence of land cover on surrounding microclimate conditions, effectively modeling the relationship between surface characteristics and temperature dynamics.

\begin{figure}[H]
  \centering
  \captionsetup{list=yes}
  \includegraphics[width=0.58\linewidth]{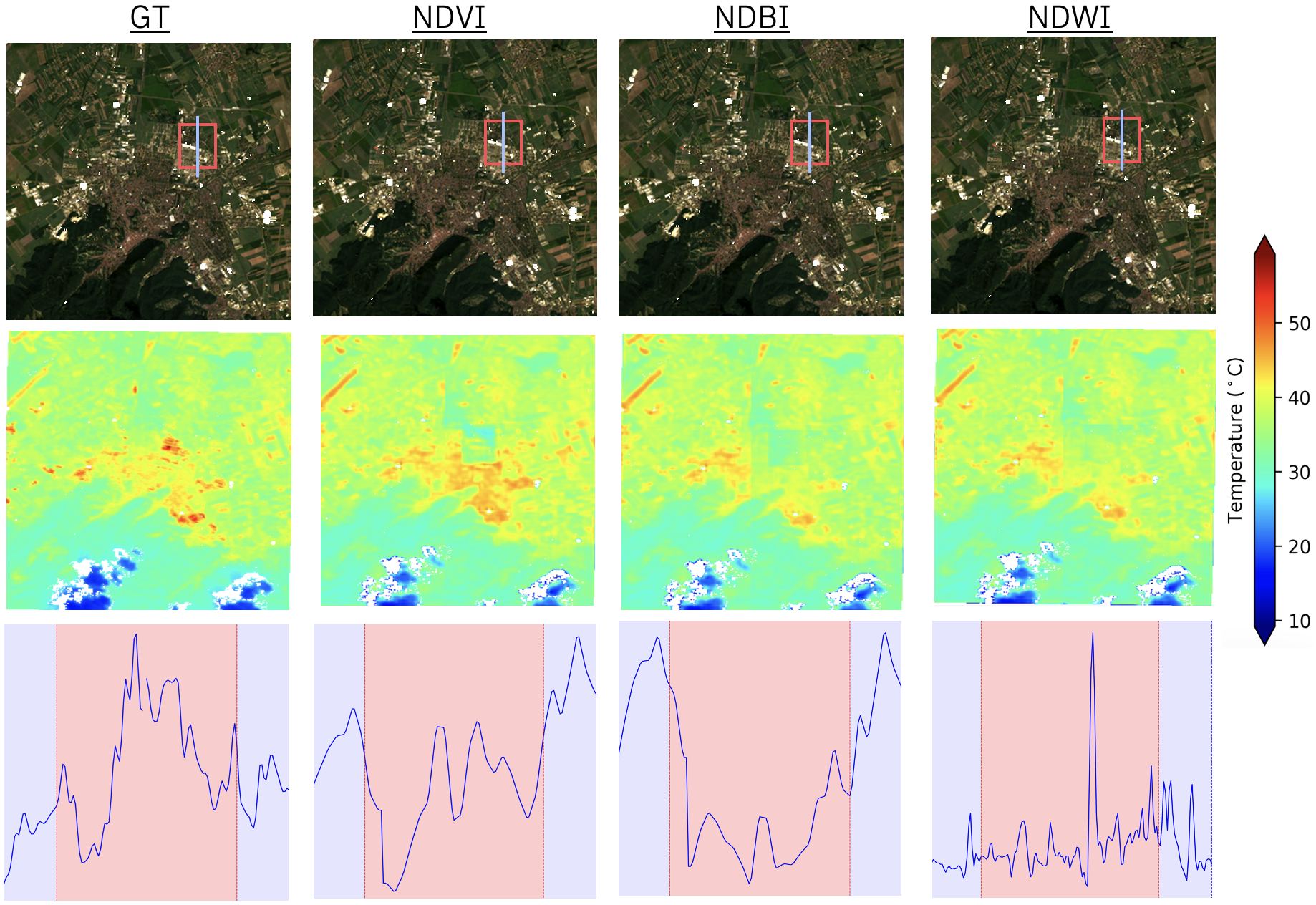} 
    \caption[Effect of spectral index band modification on inpainting results]{Inpainting results showing the effect of modifying NDVI, NDBI, and NDWI input bands, compared to ground truth.}
  \label{fig:nd-indices-swap}
\end{figure}

Another inpainting approach, described in Section~\ref{section:experiments}, modifies input bands to reflect changes in spectral indices such as NDVI, NDBI, and NDWI. The resulting predictions indicate how well the model captures underlying physical relationships. Figure~\ref{fig:nd-indices-swap} shows the model’s response to these modifications.

Consistent with Figure~\ref{fig:pixel-inpainting}, the red bounding box highlights the modified region, while the blue line marks the vertical sampling path, with corresponding temperature profiles shown below each LST image. Blue segments represent unmodified areas, while red segments indicate temperatures within the modified region.

The results suggest that the model only partially captures the physical significance of spectral indices and may rely more heavily on the RGB bands than on the spectral channels included in the satellite imagery. While NDVI behaves as expected, changes in NDBI and NDWI lead to temperature responses that are inconsistent with known physical relationships. This may stem from the masked autoencoding strategy, which prioritizes scene reconstruction over learning physical dependencies.

\section{Conclusion}
\subsection{Findings}
This study fine-tuned a GFM to evaluate its ability to predict UHI effects under future climate scenarios and evolving urban landscapes. Trained on satellite imagery and climate data, the model achieved time-averaged, pixel-wise downscaling errors below $1.74^\circ\mathrm{C}$ and closely reproduced spatial and temporal patterns from ground truth observations, demonstrating an extrapolation capacity of up to $3.62^\circ\mathrm{C}$. Its performance reflects the benefit of pre-learned features from the Prithvi model, such as vegetation and building patterns, combined with temperature fields encoding spatiotemporal variability.
LULC inputs further clarified where the model performed best in capturing UHI intensity. Simulated swaps between urban and forest pixels led to realistic shifts in LST, suggesting that the model can capture surface–microclimate interactions. 
These findings align with broader evidence that fine-tuned foundation models generalize well across tasks. As noted by \textcite{mai2023opportunitieschallengesfoundationmodels}, the combination of large-scale pretraining and task-specific fine-tuning, enabled by transfer learning, allows foundation models to adapt effectively to local urban contexts. However, the model’s limited responsiveness to complex spectral indices points to a need for improved interpretability and physical realism.

\subsection{Limitations} 
While GFMs offer strong generalization, their predictive accuracy often falls short compared to deep learning models trained for individual cities. Recent studies by \textcite{baduguPredictingLandSurface2024} or \textcite{guptaUrbanLandSurface2024} report significantly lower RMSE values using localized approaches. In contrast, GFMs prioritize broad applicability and may need further refinement to reach similar precision \cite{10641750}.

Methodological limitations also affect reliability. The model was fine-tuned on ERA5 but applied to EURO-CORDEX data, which introduces uncertainty due to differences in resolution and spectral characteristics. Beyond data inconsistencies, the model’s transferability remains untested, as fine-tuning was limited to Brașov without evaluation in other regions. Furthermore, the mitigation scenarios assumed static conditions and did not account for dynamic factors such as albedo, humidity, and human activity, all of which influence local temperatures.

\subsection{Outlook}
Further enhancements may include exploring alternative architectures, domain-specific constraints, and multimodal inputs to improve accuracy and interpretability.
Recent work suggests that integrating machine learning with physics-informed models or prior knowledge can bridge gaps between data-driven and physical approaches \cite{yangDatadrivenPhysicsinformedNeural2024}. 
Additionally, more advanced inpainting methods could improve the realism of simulated land cover changes by incorporating typical urban growth patterns and physical constraints, such as elevation limits on vegetation placement.

\printbibliography
\end{document}